\documentclass[twoside]{article}

\usepackage[accepted]{aistats2019}
\usepackage[round]{natbib}

\usepackage{amsmath}
\usepackage{graphicx}
\usepackage{amssymb}

\newcommand{\w}{\mathbf{w}}
\newcommand{\walt}{\mathbf{\tilde{w}}}
\newcommand{\ncond}{2n}
\newcommand{\suffstat}{\tilde{q}}
\newcommand{\Rb}{\mathbb{R}}

\newcommand{\s}{\mathbf{s}}
\newcommand{\xaug}{\tilde{\x}}

\newcommand{\cb}{\mathbf{c}}

\newcommand{\qmarg}{\bar{q}}
\newcommand{\qmarr}{q_0}
\newcommand{\x}{\mathbf{x}}

\newcommand{\z}{\mathbf{z}}
\newcommand{\y}{\mathbf{y}}

\newcommand{\h}{\mathbf{h}}
\newcommand{\uu}{\mathbf{u}}

\newcommand{\vb}{\mathbf{v}}

\renewcommand{\u}{\mathbf{u}}
\newcommand{\f}{\mathbf{f}}
\newcommand{\g}{\mathbf{g}}

\newcommand{\J}{\mathbf{J}}

\newcommand{\M}{\mathbf{M}}

\renewcommand{\L}{\mathbf{L}}
\newcommand{\A}{\mathbf{A}}
\newcommand{\V}{\mathbf{V}}
\newcommand{\W}{\mathbf{W}}

\newcommand{\vv}{\mathbf{v}}

\newtheorem{theorem}{Theorem}

\newtheorem{definition}{Definition}

\renewcommand{\a}{\mathbf{a}}
\renewcommand{\b}{\mathbf{b}}

\begin{document}

%

%

\twocolumn[

\runningtitle{Nonlinear ICA Using Auxiliary Variables}
\aistatstitle{Nonlinear ICA Using Auxiliary Variables\\ and Generalized Contrastive Learning}



\aistatsauthor{Aapo Hyv\"{a}rinen $^{1,2}$ \And Hiroaki Sasaki $^{3,1}$  \And Richard E.\ Turner $^{4}$}
\runningauthor{Aapo Hyv\"{a}rinen, Hiroaki Sasaki, Richard E.\ Turner}

\aistatsaddress{
\parbox{3.2cm}{\begin{center}$^{1}$ The Gatsby Unit\\UCL, UK\end{center}}
\parbox{4.2cm}{\begin{center}$^{2}$ Dept. of CS and HIIT\\Univ.\ Helsinki, Finland\end{center}}
\parbox{3.2cm}{\begin{center}$^{3}$Div.\ of Info.\ Sci.\\NAIST, Japan\end{center}}
\parbox{3.7cm}{\begin{center}$^{4}$ Univ.\ Cambridge \&\\Microsoft Research, UK \end{center}}
}

]

%

\begin{abstract}
Nonlinear ICA is a fundamental problem for unsupervised  representation learning, emphasizing the capacity to recover the  underlying latent variables generating the data (i.e.,  identifiability).  Recently, the very first identifiability proofs  for nonlinear ICA have been proposed, leveraging the temporal  structure of the independent components.  Here, we propose a general  framework for nonlinear ICA, which, as a special case, can make use  of temporal structure. It is based on augmenting the data by an  auxiliary variable, such as the time index, the history of the time  series, or any other available information. We propose to learn  nonlinear ICA by discriminating between true augmented data, or data  in which the auxiliary variable has been randomized.  This enables  the framework to be implemented algorithmically through logistic  regression, possibly in a neural network.  We provide a  comprehensive proof of the identifiability of the model as well as  the consistency of our estimation method. The approach not only  provides a general theoretical framework combining and generalizing  previously proposed nonlinear ICA models and algorithms, but also brings practical  advantages.

\end{abstract}

\section{INTRODUCTION}

Nonlinear ICA is a fundamental problem in unsupervised learning which has attracted a considerable amount of attention recently. It promises a principled approach to representation learning, for example using deep neural networks.
Nonlinear ICA attempts to find nonlinear components, or features, in multidimensional data, so that they correspond to a well-defined generative model \citep{Hyvabook,Jutten10}. The essential difference to most methods for unsupervised representation learning is that the approach starts by defining a generative model in which the original latent variables can be recovered, i.e.\ the model is identifiable by design. 

Denote an observed $n$-dimensional random vector by $\x=(x_1,\ldots,x_n)$. We assume it is generated using  $n$ independent latent variables called independent components, $s_i$. A straightforward definition of the nonlinear ICA problem is to assume that the observed data is an arbitrary (but smooth and invertible) transformation $\f$ of the latent variables $\s=(s_1,\ldots,s_n)$ as
\begin{equation} \label{genmix}
\x=\f(\s)
\end{equation}
The goal is then to recover the inverse function $\f^{-1}$ as well as the independent components $s_i$ based on observations of $\x$ alone.

Research in nonlinear ICA has been hampered by the fact that such simple approaches to nonlinear ICA are not identifiable, in stark contrast to the linear ICA case. In particular, if the observed data $\x$ are obtained as i.i.d.\ samples, i.e.\ there is no temporal or similar structure in the data, the model is seriously unidentifiable \citep{Hyva99NN}, although attempts have been made to estimate it nevertheless, often by minimizing the mutual information of outputs of a neural network \citep{Deco95,Tan01,Almeida03,brakel2017learning,hjelm2018learning}.
This is a major problem since in fact most of the utility of linear ICA rests on the fact that the model is identifiable, or---in alternative terminology---the ``sources can be separated''. Proving the identifiability of linear ICA \citep{Comon94} was a great advance on the classical theory of factor analysis, where an orthogonal factor rotation could not be identified.

Fortunately, a solution to non-identifiability in nonlinear ICA can be found by utilizing temporal structure in the data \citep{Harmeling03,Sprekeler14,Hyva16NIPS,Hyva17AISTATS}. In recent work, various identifiability conditions have been proposed, assuming that the independent components are actually time series and have autocorrelations \citep{Sprekeler14}, general non-Gaussian temporal dependencies \citep{Hyva17AISTATS}, or non-stationarities \citep{Hyva16NIPS}. These generalize earlier identifiability conditions for linear ICA with temporal structure \citep{Belo97,Pham01}.


Meanwhile, recent work in computer vision has successfully proposed ``self-supervised'' feature extraction methods from a purely heuristic perspective. The method by \citet{misra2016shuffle} is quite similar to the nonlinear ICA by \citet{Hyva17AISTATS}, while \citet{oord2018representation} proposed a method related to \citet{Hyva16NIPS}---see also further self-supervised methods by \citet{noroozi2016unsupervised,larsson2017colorization}. These approaches have allowed unsupervised data to be leveraged for supervised tasks resulting in dramatic performance improvements, but the papers acknowledge that they lack theoretical grounding. 

Here, we propose a very general form of nonlinear ICA, based on the idea that the independent components are dependent on some additional auxiliary variable, while being conditionally mutually independent given the auxiliary variable. This unifies and generalizes the methods by \citet{Harmeling03,Sprekeler14,Hyva16NIPS,Hyva17AISTATS}, giving  a general framework where it is not necessary to specifically have a temporal (or even spatial) structure in the data. 
We prove exact identifiability conditions for the new framework, and show how it extends previous conditions, both from the viewpoint of theory and practice.
In particular, our theory establishes mathematical principles underlying an important strand of self-supervised approaches by \citet{arandjelovic2017look} and \citet{korbar2018co}, showing that under certain conditions, they will extract the underlying latent variables from the data.
 We further provide a practical algorithm for estimating the model using the idea of contrastive learning \citep{Gutmann12JMLR,Hyva16NIPS,Hyva17AISTATS}, and prove its consistency.

\section{BACKGROUND}

We start by giving some background on nonlinear ICA theory. We explain the central problem of unidentifiability of nonlinear ICA, and discuss some recently proposed solutions using time structure.


A straightforward generalization of ICA to the nonlinear case would assume, as pointed out above, a mixing model (\ref{genmix}) with mutually independent latent variables $s_i$, and a general nonlinear mixing function $\f$, only assumed to be invertible and smooth. 
Now, if we further assume that the observations of $\x$ are independent and identically distributed (i.i.d.), the model is seriously unidentifiable. A well-known result, see e.g.\  \citet{Hyva99NN}, shows how to construct a function $g$ such that for any two random variables $x_1$ and $x_2$, the function $g(x_1,x_2)$ is independent of $x_1$. This leads to the absurd case where based on independence alone, we could consider any of the observed variables an independent component. 

Nor can we get any new information based on the non-Gaussianity of the variables, like in linear ICA, because we can trivially create a point-wise transformation  $f(x_i)$ to have \textit{any} marginal distribution by well-known theory (compounding the inverse cdf of the target distribution and the cdf of $x_i$).

One possibility to obtain identifiability is to restrict the nonlinearity $\f$. However, very few results are available in that direction, and usually based on very restrictive conditions, such as adding scalar nonlinearities to a linear mixing  \citep{Taleb99}.


A more promising direction is to relax the assumption of i.i.d.\ sampling. A fundamental case  is to consider time series, and the \textit{temporal structure} of independent components. Thus, we assume
\begin{equation} \label{genmixtemp}
\x(t)=\f(\s(t))
\end{equation}
where $t$ is the time index.
As a first attempt, we can assume that the sources $s_i(t)$ have non-zero \textit{autocorrelations}, which has a long history in the linear case \citep{Tong91,Belo97}. \citet{Harmeling03} proposed that we could try to find nonlinear transformations which are maximally uncorrelated even over time lags and after nonlinear scalar transformations. \citet{Sprekeler14} showed that a closely related method enables separation of sources if they all have distinct autocorrelations functions. This constitutes probably the first identifiability proof for nonlinear ICA with general nonlinearities. However, it suffers from the restrictive condition that the sources must have different statistical properties, which is rather unrealistic in many cases.


An alternative framework was proposed by \cite{Hyva17AISTATS}, where it was first heuristically proposed to transform the problem into a classification problem between two data sets, one constructed by concatenating real data points by taking a time window of data, and the other by a randomized (permuted) concatenation. In other words, we define two data sets such as:
\begin{align}
\xaug(t)=(\x(t),\x(t-1))   \ \text{    vs.    } \ 
\xaug^*(t)=(\x(t),\x(t^*))
\end{align}
with a random time index $t^*$. We then train a neural network to discriminate between these two new data sets.
Such Permutation-Contrastive Learning (PCL) \citep{Hyva17AISTATS} was shown to estimate independent components in the hidden layer, even if they have identical distributions, assuming they have \textit{temporal dependencies} which are, loosely speaking, non-Gaussian enough. In the case of Gaussian sources, PCL estimates the sources under the same conditions as the method by \citet{Sprekeler14}. Thus, the PCL theory proves a stronger version of identifiability based on temporally dependent sources.


Another form of temporal structure that has been previously used in the case of linear ICA is \textit{nonstationarity} \citep{Matsuoka95}. 
This principle was extended to the nonlinear case by \citet{Hyva16NIPS}. 
The starting point was a heuristic principle where the time series is divided into a large number of segments. Then, a neural network is trained by multinomial regression so that each data point (i.e.\ time point) is assigned an artificially defined label given by the index of the time segment to which it belongs. Intuitively speaking, one would expect that the hidden layers of the neural network must learn to represent nonstationarity, since nonstationarity is nothing else than the differences between the distributions of the time segments.
The ensuing method, Time-Contrastive Learning (TCL), was actually shown to enable estimation of a nonlinear ICA model where the independent components are assumed to be nonstationary,  at the same time constituting another identifiability proof. Note that nonstationarity and temporal dependencies are two completely different properties which do not imply each other in any way.

\section{NONLINEAR ICA USING AUXILIARY VARIABLES}

Next, we propose our general framework for nonlinear ICA, as well as a practical estimation algorithm.

\subsection{Definition of generative model}
Assume the general $\Rb^n\rightarrow \Rb^n$ mixing model in (\ref{genmix})
where the mixing function $\f$ is only assumed invertible and smooth (in the sense of having continuous second derivatives, and the same for its inverse). We emphasize the point that we do not restrict the function $\f$ to any particular functional form. It can be modelled by a general neural network, since even the assumption of invertibility usually (empirically) seems to hold for the $\f$ estimated by the methods developed here, even without enforcing it.

The key idea here is that we further assume that each $s_i$ is statistically dependent on some fully-observed $m$-dimensional random variable $\uu$, but conditionally independent of the other $s_j$:
\begin{equation} \label{genprior}
\log p(\s|\uu)= \sum_{i=1}^n q_i(s_i,\uu)
\end{equation}
for some functions $q_i$. 
First, to see how this generalizes previous work on nonlinear ICA using time structure, we note that the auxiliary variable $\uu$ could be
the past of the component in the time series, giving rise to temporally dependent components as in permutation-contrastive learning or PCL \citep{Hyva17AISTATS} and the earlier methods by \citet{Harmeling03,Sprekeler14}. 
Alternatively, $\uu$ could be the time index $t$ itself in a time series, or the index of a time segment, leading to nonstationary components as in time-contrastive learning or TCL \citep{Hyva16NIPS}. These connections will be considered in more detail below.

Thus, we obtain a unification of the separation principles of temporal dependencies and non-stationarity. This is remarkable since these principles are well-known in the linear ICA literature, but they have been considered as two distinct principles \citep{Cardoso01,Hyvabook}.

Furthermore, we can define $\uu$ in completely new ways. In the case where each observation of $\x$ is an image or an image patch, a rather obvious generalization of TCL would be to assume $\uu$ is the pixel index, or any similar spatial index, thus giving rise to nonlinear representation learning by the $\s$. In a visual feature extraction task, $\x$ could be images and $\uu$ related audio or text \citep{arandjelovic2017look}.
Moreover, $\uu$ could be a class label, giving rise to something more related to conventional representation learning by a supervised neural network (e.g.\ ImageNet), but now connected to the theory of nonlinear ICA and identifiability (this is also considered in detail below). In a neuroscience context, $\x$ could be brain imaging data, and the $\uu$ could be some quantity related to the stimuli in the experiment. Furthermore, $\uu$ could be some combination of some of the above, thus providing a very general method. 
The appropriate definition of $\uu$ obviously depends on the application domain, and the list above is by no means exhaustive.

It should be noted that the conditional independence does not imply that the $s_i$ would be marginally independent. If $\uu$ affects the distributions of the $s_i$ somehow independently (intuitively speaking), the $s_i$ are likely to be marginally independent. This would be case, for example, if each $q_i$ is of the from $q_i(s_i,u_i)$, that is, each source has one auxiliary variable which is not shared with the other sources, and the $u_i$ are independent of each other. Thus, the formulation above is actually generalizing the ordinary independence in ICA to some extent.

\subsection{Learning algorithm}
To estimate our nonlinear ICA model, we propose a general form of contrastive learning, inspired by the idea of transforming unsupervised learning to supervised learning previously explored by \citet{Gutmann12JMLR,Goodfellow14,gutmann2014likelihood}. More specifically, we use the idea of discriminating between a real data set and some randomized version of it, as used in PCL. Thus we define two datasets
\begin{align}
\xaug=(\x,\uu)  \  \text{ vs. } \
\xaug^*=(\x,\uu^*) \label{xtildestar}
\end{align}
where $\uu^*$ is a random value from the distribution of the $\uu$, but independent of $\x$, created in practice by random permutation of the empirical sample of the $\uu$. We learn a nonlinear logistic regression system (e.g.\ a neural network) using a regression function of the form
\begin{equation} \label{reggen}
r(\x,\u)= \sum_{i=1}^n \psi_i(h_i(\x),\uu)
\end{equation}
which then gives the posterior probability of the first class as $1/(1+\exp(-r(\x,\uu))$. Here, the scalar features $h_i$ would typically be computed by hidden units in a neural network. Universal approximation capacity \citep{Hornik} is assumed for the models of $h_i$ and $\psi_i$. This is a variant of the ``contrastive learning'' approach to nonlinear ICA  \citep{Hyva16NIPS,Hyva17AISTATS}, and we will see below that it in fact unifies and generalizes those earlier results. 

\section{THEORETICAL ANALYSIS}

In this section, we give exact conditions for the convergence (consistency) of our learning algorithm, which also leads to constructive proofs of identifiability of our nonlinear ICA model with auxiliary variables. It turns out we have two cases that need to be considered separately, based on the property of conditional exponentiality.

\subsection{Definition of  conditional exponentiality}

We start by a basic definition describing distributions which are in some sense pathological in our theory.
\begin{definition} \label{monoexpdef}
A random variable (independent component) $s_i$ is conditionally exponential of order $k$ given random vector $\uu$ if its conditional pdf can be given in the form
\begin{equation} \label{expdef}
  p(s_i|\uu)=\frac{Q_i(s_i)}{Z_i(\uu)}\exp[\sum_{j=1}^k \suffstat_{ij}(s_i)\lambda_{ij}(\uu)]
\end{equation}
almost everywhere in the support of $\uu$, with $\suffstat_{ij}$, $\lambda_{ij}$, $Q_i$, and $Z_i$  scalar-valued  functions. The sufficient statistics $\suffstat_{ij}$ are assumed linearly independent (over $j$, for each fixed $i$). 
\end{definition}
This definition is a simple variant of the conventional theory of exponential families, adding conditioning by $\uu$ which comes through the parameters only. 

As a simple illustration, consider a (stationary) Gaussian time series as $s_i$, and define $\uu$ as the past of the time series. The past of the time series can be compressed in a single statistic $\lambda(\uu)$ which essentially gives the conditional expectation of $s_i$. Thus, models of independent components using Gaussian autocorrelations lead to the conditionally exponential case, of order $k=1$. 
As is well-known, the basic theory of linear ICA relies heavily on non-Gaussianity, the intuitive idea being that the Gaussian distribution is too ``simple'' to support identifiability. Here, we see a reflection of the same idea. Note also that if $s_i$ and $\uu$ are independent, $s_i$ is conditionally exponential, since then we simply set $k=1, \suffstat_{i1}\equiv 0$.

In the following, we analyse our algorithm separately for the general case, and for conditionally exponential independent components of low order $k$. The fundamental result is that for sufficiently complex source distributions, the independent components are estimated up to component-wise nonlinear transformations; if the data comes from an exponential family of low order, there is an additional linear transformation that remains to be determined (by linear ICA, for example).

\subsection{Theory for general case}

First, we consider the much more general case of distributions which are not ``pathological''.
Our main theorem, proven in Supplementary Material~\ref{proof1}, is as follows:
\begin{theorem}\label{Th1}
Assume
\begin{enumerate}
\item The observed data follows the nonlinear ICA model with auxiliary variables in Eqs.~(\ref{genmix},\ref{genprior}). 
\item The conditional log-pdf $q_i$ in (\ref{genprior}) is sufficiently smooth as a function of $s_i$, for any fixed $\uu$. 
\item \textbf{[Assumption of Variability]} For any $\y\in \Rb^n$, there exist $\ncond+1$ values for $\uu$, denoted by $\uu_j,j=0...\ncond$ such that the $\ncond$ vectors in $\Rb^{2n}$ given by
\begin{multline}
\left(\w(\y,\uu_1)-\w(\y,\uu_0)),(\w(\y,\uu_2)-\w(\y,\uu_0)),\right. \\ \left. ...,(\w(\y,\uu_{\ncond})-\w(\y,\uu_0)\right)
\end{multline}
with
\begin{multline} \label{wdef}
\w(\y,\uu)=\left(\frac{\partial q_1(y_1,\uu)}{ \partial y_1},\ldots,\frac{\partial q_n(y_n,\uu)}{ \partial y_n},  \right. \\ \frac{\partial^2 q_1(y_1,\uu)}{ \partial y_1^2},   \left. \ldots,\frac{\partial^2 q_n(y_n,\uu)}{ \partial y_n^2}\right)
\end{multline}
are linearly independent.  \label{assvar}
\item We train some nonlinear logistic regression system with universal approximation capability  to discriminate between $\xaug$ and $\xaug^*$ in (\ref{xtildestar}) with regression function in (\ref{reggen}).
\item In the regression function in Eq.~(\ref{reggen}),  we constrain $\h=(h_1,...,h_n)$ to be invertible, as well as smooth, and constrain the inverse to be smooth as well.
\end{enumerate}
Then, in the limit of infinite data, $\h$ in the regression function provides a consistent estimator of demixing in the nonlinear ICA model: The functions (hidden units) $h_i(\x)$ give the independent components, up to scalar (component-wise) invertible transformations.
\end{theorem}
Essentially, the Theorem shows that under mostly weak assumptions, including invertibility of $\h$ and  smoothness of the pdfs, and of course independence of the components, our learning system will recover the independent components given an infinite amount of data. Thus, we also obtain a constructive identifiability proof of our new, general nonlinear ICA model.

Among the assumptions above, the only one which cannot considered weak or natural is clearly Assumption of Variability (\#\ref{assvar}), which is central in the our developments. It is basically saying that the auxiliary variable must have a sufficiently strong and diverse effect on the distributions of the independent components. To further understand this condition, we give the following Theorem, proven in Supplementary Material~\ref{proofnew}:
\begin{theorem}\label{Thnew}
Assume the independent components are conditionally exponential given $\uu$, with the same order $k$ for all components.
Then, 
\begin{enumerate}
\item If $k=1$, the Assumption of Variability cannot hold. 
\item Assume $k>1$ and for each component $s_i$, the vectors $(\frac{\partial \suffstat_{ij}(s_i,\uu)}{ \partial s_i}, \frac{\partial^2 \suffstat_{ij}(s_i,\uu)}{ \partial s_i^2})$ are not all proportional to each other for different $j=1,\ldots,k$, for $\s$ almost everywhere. Then,
the Assumption of Variability holds almost surely if the $\lambda$'s are statistically independent and follow a distribution whose support has non-zero measure.
\end{enumerate}
\end{theorem}
Loosely speaking, the Assumption of Variability holds if the sources, or rather their modulation by $\uu$, is not ``too simple'', which is here quantified as the order of the exponential family from which the $s_i$ are generated. Furthermore, for the second condition of Theorem~\ref{Thnew} to hold, the sufficient statistics cannot be linear (which would lead to zero second derivatives), thus excluding the Gaussian scale-location family as too simple as well. (See Supplementary Material \ref{ThApp.sec} for an alternative formulation of the assumption.)

Another non-trivial assumption in Theorem~\ref{Th1} is the invertibility of $\h$.  It is hoped that the constraint of invertibility is only necessary to have a rigorous theory, and not necessary in any practical implementation. Our simulations below, as well as our next Theorem, seem to back up this conjecture to some extent.

\subsection{Theory for conditionally exponential case}

The theory above excluded the conditionally exponential case of order one (Theorem~\ref{Thnew}). This is a bit curious since it is actually the main model considered in TCL \citep{Hyva16NIPS}. In fact, the exponential family model of nonstationarities in that work is nothing other than a special case of our ``conditionally exponential'' family of distributions; we will consider the connection in detail in the next section.

There is actually a fundamental difference between Theorem~\ref{Th1} above and the TCL theory in \citep{Hyva16NIPS}. In TCL, and in contrast to our current results, a linear indeterminacy remains---but the  TCL theory never showed that such an indeterminacy is a property of the model and not only of the particular TCL algorithm employed by \citet{Hyva16NIPS}. 





Next, we construct a theory for conditionally exponential families adapting our current framework, and indeed, we see the same kind of linear indeterminacy as in TCL appear.
We give the result for general $k$, although the case $k=1$ is mainly of interest:
\begin{theorem} \label{Th2}
Assume
\begin{enumerate}
\item The data follows the nonlinear ICA model with auxiliary variables in  Eqs.~(\ref{genmix},\ref{genprior}). 
\item Each $s_i$ is conditionally exponential given $\uu$ (Def.~\ref{monoexpdef}). 
\item There exist $nk+1$ points $\uu_0,\ldots,\uu_{nk}$ such that the matrix of size $nk\times nk$
\begin{equation}
\bar{\L}=
\begin{pmatrix}
\lambda_{11}(\uu_1) - \lambda_{11}(\uu_0), ..., \lambda_{11}(\uu_{nk}) - \lambda_{11}(\uu_0)\\
\vdots\\
\lambda_{nk}(\uu_1) - \lambda_{nk}(\uu_0), ..., \lambda_{nk}(\uu_{nk}) - \lambda_{nk}(\uu_0)\\
\end{pmatrix}
\end{equation}
 is invertible (here, the rows corresponds to all the $nk$ possible subscript pairs for $\lambda$).
\item We train a nonlinear logistic regression system with universal approximation capability to discriminate between $\xaug$ and $\xaug^*$ in (\ref{xtildestar}) with regression function in (\ref{reggen}).
\end{enumerate}
Then, 
\begin{enumerate}
\item The optimal regression function can be expressed in the form
\begin{equation}  \label{rmonoexp}
r(\x,\uu)=\tilde{\h}(\x)^T \vv(\uu) + a(\x) + b(\uu)
\end{equation} 
for some functions $\vv:\Rb^{m}\rightarrow\Rb^{nk}$, $\tilde{\h}:\Rb^n\rightarrow \Rb^{nk}$ and two scalar-valued functions $a,b$. 
\item In the limit of infinite data, $\tilde{\h}(\x)$ provides a consistent estimator of the nonlinear ICA model, up to a linear transformation of point-wise scalar (not necessarily invertible) functions of the independent components. The point-wise nonlinearities are given by the sufficient statistics $\suffstat_i$. In other words,
\begin{multline} \label{monoexptheorem}
\begin{pmatrix}
\suffstat_{11}(s_1),
\suffstat_{12}(s_1),
\hdots,
\suffstat_{21}(s_2),
\hdots,
\suffstat_{nk}(s_n)
\end{pmatrix}^T
\\= \A \tilde{\h}(\x) - \mathbf{c}
\end{multline}
for some unknown matrix $\A$ and an unknown vector $\mathbf{c}$.
\end{enumerate}
\end{theorem}
The proof, found in Supplementary Material~\ref{proof2}, is quite similar to the proof of Theorem~1 by \citet{Hyva16NIPS}. Although the statistical assumptions made here are different, the very goal of modelling exponential sources by logistic regression means the same linear indeterminacy appears, based on the linearity of the log-pdf in exponential families. 




\section{DIFFERENT DEFINITIONS OF AUXILIARY VARIABLES}

Next, we consider different possible definitions of the auxiliary variable, and show some exact connections and generalization of previous work. 

First, we want to emphasize that any arbitrary definition of $\uu$ is not possible since many definitions are likely to violate the central assumption of conditional independence of the components. For example, one might be tempted to choose a $\u$ which is some deterministic function of $\x$; in Supplementary Material~\ref{ufx.sec} we give a simple example showing how this violates the conditional independence.

\subsection{Using time as auxiliary variable}

A real practical utility of the new framework can be seen in the case of nonstationary data. 
Assume we observe a time series $\x(t)$ as in Eq.~(\ref{genmixtemp}).
Assume the $n$ independent components are nonstationary, with densities $p(s_i|t)$. 
For analysing such nonstationary data in our framework, define $\x=\x(t)$ and $\uu=t$. We can easily consider the time index as a random variable, observed for each data point, and coming from a uniform distribution. 
Thus, we create two new datasets by augmenting the data by adding the time index:
\begin{align}
\xaug=(\x(t),t) \  \text{ vs. } \ 
\xaug^*=(\x(t),t^*)
\end{align}
We  analyse the nonstationary structure of the data by learning to discriminate between $\xaug$ and $\xaug^*$ by logistic regression. Directly applying the general theory above, we define the regression function to have the following form:
\begin{equation} \label{reg}
r(\x,t)= \sum_i \psi_i(h_i(\x),t)
\end{equation}
where each $\psi_i$ is $\Rb^2\rightarrow\Rb$. 
Intuitively, this means that the nonstationarity is separately modelled for each component, with no interactions.

Theorems~\ref{Th1} and \ref{Th2} above give exact conditions for the consistency of such a method. This provides an alternative way of estimating the nonstationary nonlinear ICA model proposed in \citep{Hyva16NIPS} as a target for the  TCL method. 

A practical advantage is that if the assumptions of Theorem~\ref{Th1} hold, the method actually captures the independent components directly: There is no indeterminacy of a linear transformation unlike in TCL. Nor is there any nonlinear non-invertible transformation (e.g.\ squaring) as in TCL, although this may come at the price of constraining $\h$ to be invertible. The Assumption of Variability in Theorem~\ref{Th1} is quite comparable to the corresponding full rank condition in the convergence theory of TCL.
Another advantage of our new method is that there is no need to segment the data, although 
in our simulations below we found that segmentation is computationally very useful.
From a theoretical perspective, the current theory in Theorem~\ref{Th1} is also much more general than the TCL theory since no assumption of an exponential family is needed --- ``too simple'' exponential families are in fact considered separately in Theorem~\ref{Th2}. 

\subsection{Using history as auxiliary variables} \label{pclcomp.sec}

Next, we consider the theory in the case where $\uu$ is the history of each variable. 
For the purposes of our present theory, we define $\x=\x(t)$ and $\uu=\x(t-1)$ based on a time-series model in (\ref{genmixtemp}). So, the nonlinear ICA model in Eqs.~(\ref{genmix}, \ref{genprior}) holds. Note that here, it does not make any difference if we use the past of $\x$ or of $\h(\x)$ as $\uu$ since they are invertible functions of each other. 
Each component follows a distribution
\begin{equation}
q_i(s_i,\uu)= q_i(s_i(t),s_i(t-1))
\end{equation}
This model is the same as in PCL \citep{Hyva17AISTATS}, and in fact PCL is thus a special case of the discrimination problem we formulated in this paper.
Likewise, the restriction of the regression function in (\ref{reggen}) is very similar to the form imposed in Eq.~(12) of \citep{Hyva17AISTATS}.
Thus, essentially, Theorem~\ref{Th1} above provides  an alternative identifiability proof of the model in \citep{Hyva17AISTATS}, with quite similar constraints.
See Supplementary Material~\ref{pclcompappendix.sec} for a detailed discussion on the connection. 
Our goal here is thus not to sharpen the analysis of \citep{Hyva17AISTATS}, but merely to show that that model falls into the present framework with minimal modification.

\subsection{Combining time and history} \label{combine.sec}

Another generalization of previously published theory which could be of great interest in practice is to combine the nonstationarity-based model in TCL \citep{Hyva16NIPS} with the temporal dependencies model in PCL \citep{Hyva17AISTATS}.
Clearly, we can combine these two by defining $\uu=(\x(t-1),t)$, and thus discriminating between
\begin{align}
\xaug(t)=(\x(t),\x(t-1),t) \text{ vs. }
\xaug^*(t)=(\x(t),\x(t^*-1),t^*) \nonumber
\end{align}
with a random time index $t^*$, and accordingly defining the regression function as
\begin{equation} \nonumber 
r_{\text{comb}}(\x(t),\x(t-1),t)=\sum_{i=1}^n \psi_i(h_i(\x(t)),h_i(\x(t-1)),t)  
\end{equation}
Such a method now has the potential of using both nonstationarity and temporal dependencies for nonlinear ICA. Thus, there is no need to choose which method to use, since this combined method uses both properties. (See Supplementary Material~\ref{combineappendix.sec} for an alternative formulation.)

\subsection{Using class label as auxiliary variable} \label{classlabel.sec}

Finally, we consider the very interesting case where the data includes class labels as in a classical supervised setting, and we use them as the auxiliary variable. Let us note that the existence of labels does not mean a nonlinear ICA model is not interesting, because our interest might not be in classifying the data using these labels, but rather in understanding the structure of the data, or possibly, finding useful features for classification using some other labels as in transfer learning. 
In particular, with scientific data, the main goal is usually to understand its structure; if the labels correspond to different treatments, or experimental conditions, the classification problem in itself may not be of great interest.
It could also be that the classes are somehow artificially created, as in TCL, and thus the whole classification problem is of secondary interest.

Formally, denote by $c\in\{1,..,k\}$ the class label with $k$ different classes. As a straight-forward application of the theory above, we learn to discriminate between 
\begin{align}
\xaug=(\x,c)  \ \text{ vs. } \ 
\xaug^*=(\x,c^*) \label{xtildestarclass}
\end{align}
where $c$ is the class label of $\x$, and $c^*$ is a randomized class label; ``one-hot'' coding of $c$ could also be used. 
Note that we could also apply the TCL method and theory on such data, simply using the $c$ as class labels instead of the time segment indices as in \citep{Hyva16NIPS}. 
Applying Theorem~\ref{Th1}, we see that, interestingly, we have no linear indeterminacy, unlike in TCL (unless the data follows a conditionally exponential source model of low rank, in which case we fall back to Theorem~\ref{Th2}.) Thus, the current theorem seems to be in some sense stronger than the TCL theory, although it is not a strict generalization.
In either case, we use the class labels to estimate independent components, thus combining supervised and unsupervised learning in an interesting, new way.


\section{SIMULATIONS}

To test the performance of the method, we applied it on non-stationary sources similar to those used in TCL. This is the case of main interest here since for temporally correlated sources, the framework gives PCL. It is not our goal to claim that the new method performs better than TCL, but rather to confirm that our new very general framework includes something similar to TCL as well.

\begin{figure}[tb]
\begin{center}
 \centering
 \raisebox{2.9cm}{\textsf{a)}}
 \includegraphics[height=3.3cm]{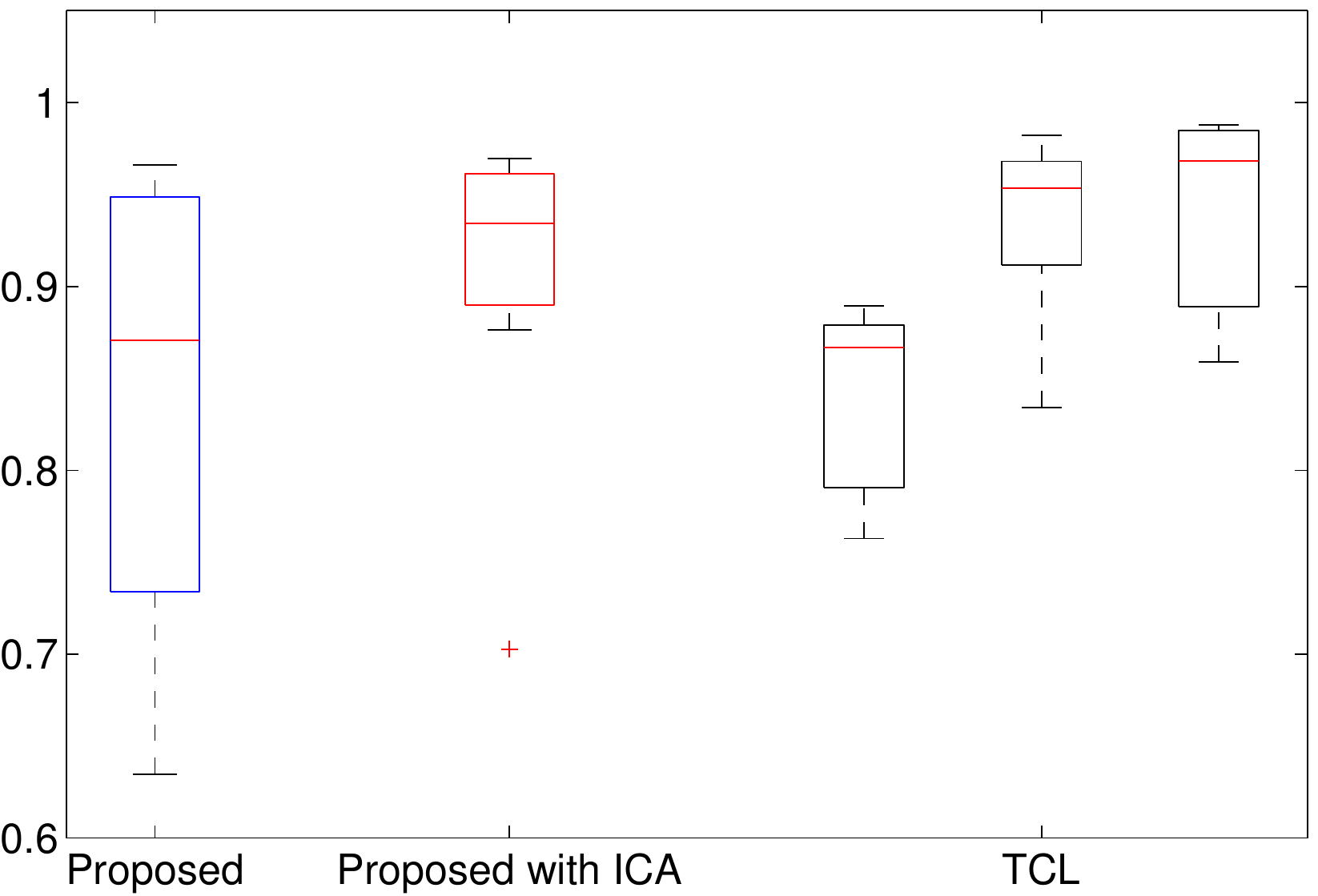} \\
 \raisebox{2.9cm}{\textsf{b)}}
 \includegraphics[height=3.3cm]{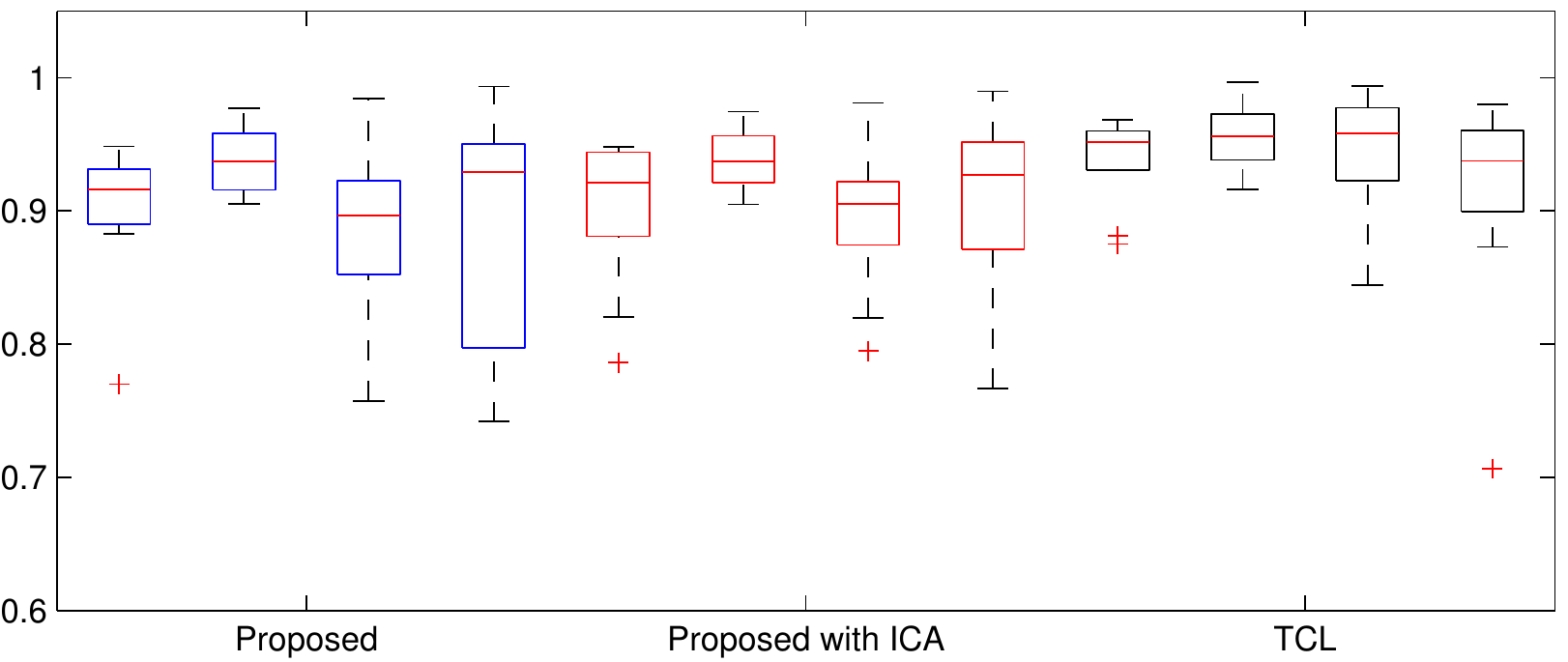}
\end{center}
 \caption{Performance measured by correlations between estimates and original quantities (see text). The non-conditionally-exponential case is given in (a) and the exponential family case in (b). 
 ``Proposed'' is taking raw outputs from neural network learned by our new method, ``Proposed with ICA'' is adding final linear ICA, ``TCL'' is time-contrastive learning (with final linear ICA) given for comparison. In a), TCL was performed with $16, 64$, and $256$ time segments.
In b), for each method, we report four cases, with $10, 50,100$, and $300$ time segments.
}
\label{Fig}
\end{figure}

First, we consider the non-conditionally-exponential case in Theorem~\ref{Th1}, where the data does not follow a conditionally exponential family, and the regression function has the general form in (\ref{reggen}). We artificially generated nonstationary sources $s_i$ on a 2D grid indexed by $\xi,\eta$ by a scale mixture model:\linebreak $s_i(\xi,\eta) = \sigma_i(\xi,\eta)\cdot
z_i(\xi,\eta)$, $i=1,\dots,n$, where $z_i$ is a standardized Laplacian variable, and the scale components $\sigma_i(\xi,\eta)$ were generated by creating Gaussian blobs in random locations to represent areas of higher variance. The number of dimensions was $5$ and the number of data points $2^{16}$.
The mixing function $\f$ was a random three-layer feedforward neural network as
in~\citep{Hyva16NIPS}.
We used the spatial index pair as $\uu:=(\xi,\eta)$.
We modelled $\h(\x)$ by a feedforward neural network with three layers: The number of units in the hidden layers was $2n$, except in the final layer where it was $n$; the nonlinearity was max-out except for the last layer where absolute values were taken; $L_2$ regularization was used to prevent overlearning.  The function $\psi_i$ was also modelled by a neural network. In contrast to the assumptions of Theorem~\ref{Th1}, no constraint related to the invertibility of $h$ was imposed.
After learning the neural network, we further applied FastICA to the estimated features (heuristically inspired by Theorem~\ref{Th2}).
Performance was evaluated by the Pearson correlation between the estimated sources and the original sources (after optimal matching and sign flipping).
The results are shown in Fig.~\ref{Fig} a). Our method has performance similar to TCL. 

Second, we considered the conditionally exponential family case as in Theorem~\ref{Th2}.
We generated nonstationary sources
 $s_i$ as above, but we
generated them as time-series, and divided the time series into equispaced segments. We used a simple random neural network to generate separate variances $\sigma_i$ inside each segment. 
The mixing function was as above.
Here, we used the index of the segment as $\uu$. This means we are also testing the applicability of using a class label as the auxiliary variable as in Section~\ref{classlabel.sec}.
We modelled $\h(\x)$ as above.
The $\vb$  and $b$ in (\ref{rmonoexp}) were modelled by constant parameter vectors inside each segment, and $a$ by another neural network.
Performance was evaluated by the Pearson correlation of the absolute values of the components, since the sign remains unresolved in this case.
The results are shown in Fig.~\ref{Fig} b). Again our method has performance similar to TCL, confirming that source separation by nonstationarity, as well as using class labels as in Section~\ref{classlabel.sec}, can be modelled in our new framework.

\section{CONCLUSION}

We introduced a new framework for nonlinear ICA. To solve the problem of non-identifiability central to nonlinear ICA theory, we assume there is an external, auxiliary variable, such that conditioning by the auxiliary variables changes the distributions of the independent components. In a time series, the auxiliary variable can correspond to the history, or the time index, thus unifying the previous frameworks \citep{Sprekeler14,Hyva16NIPS,Hyva17AISTATS} both in theory and practice. 

We gave exact conditions for identifiability, showing how the
definition of conditional exponentiality divides the problem into
two domains. Conditional exponentiality interestingly
corresponds to the simplest case of TCL theory in \citep{Hyva16NIPS}.  
In the special case of nonstationary components like in TCL, we
actually relaxed the assumption of an exponential family
model for the independent components, and removed the need to segment
the data, which may be difficult in practice; nor was there any remaining
linear mixing, unlike in TCL. This result carried over to the case where we actually have class labels available; we argued that the identifiability theory of nonlinear ICA is interesting even in such an apparently supervised learning case.
We also provided a learning algorithm based on the idea of contrastive learning  by logistic regression, and proved its consistency.

Recent work has successfully used a very similar idea for ``self-supervised'' audio-visual feature extraction from a purely heuristic perspective \citep{arandjelovic2017look,korbar2018co}, and our theory hopefully elucidates the mathematical principles underlying such methods.
 Yet, our framework is quite versatile, and the auxiliary variables can be defined in many different ways depending on the application.



\newpage


\appendix

\newpage \
\newpage

\begin{center}
{ \bf 
 Nonlinear ICA using auxiliary variables and \\generalized contrastive learning\\AISTATS 2019\\ \  \\\Large Supplementary Material 
}
\end{center}

\section{Proof of Theorem~\ref{Th1}} \label{proof1}

By well-known theory \citep{Gutmann12JMLR,Friedmanbook}, after convergence of logistic regression, with infinite data and a function approximator with universal approximation capability, the regression function will equal the difference of the log-densities in the two classes:
\begin{multline} \nonumber
\sum_{i=1}^n \psi_i(h_i(\x),\uu) = \sum_i q_i(g_i(\x),\uu) + \log p(\uu)\\ + \log |\det \J\g(\x)| - \log p_\s (\g(\x)) - \log p(\uu) -  \log |\det \J\g(\x)|
\end{multline}
where the $\log p_\s$ is the marginal log-density of the components when $\uu$ is integrated out (as pointed above, it does not need to be factorial), $\log p(\uu)$ is the marginal density of the auxiliary variables, $\g=\f^{-1}$, and the
$\J\g$ are the Jacobians of the inverse mixing---which nicely cancel out. Also, the marginals $\log p(\uu)$  cancel out here.

Now, change variables to $\y=\h(\x)$ and define $\vb(\y)=\g(\h^{-1}(\y))$, which is possible by the assumption of invertibility of $\h$. 
We then have
\begin{equation} \label{basiceqgen}
\sum_i \psi_i(y_i,\uu) = \sum_i q_i(v_i(\y),\uu) - \log p_\s(\vb(\y)) 
\end{equation}
What we need to prove is that this can be true for all $\y$ and $\uu$ only if the $v_i$ depend on only one of the $y_i$.

Denote $\qmarg(\y)= \log p_\s(\vb(\y))$.
Taking derivatives of both sides of (\ref{basiceqgen}) with respect to $y_j$, 
denoting the derivatives by a superscript as 
\begin{align}
q_i^1(s,\uu)=\partial q_i(s,\uu)/ \partial s  \\
q_i^{11}(s,\uu)=\partial^2 q_i(s,\uu)/ \partial s^2 
\end{align}
and likewise for $\psi$, and $v_i^j(\y)=\partial v_i(\y)/ \partial y_j$, we obtain
\begin{equation}
 \psi_j^1(y_j,\uu) = \sum_i q_i^1(v_i(\y),\uu) v_i^j(\y) -  \qmarg^j(\y) 
\end{equation}
Taking another derivative with respect to $y_{j'}$ with $j'\neq j$, the left-hand-side vanishes, and we have
\begin{multline}
\sum_i q_i^{11}(v_i(\y),\uu)\, v_i^j(\y)\, v_i^{j'}(\y) 
+ q_i^{1}(v_i(\y),\uu) \,v_i^{jj'}(\y)
\\-  \qmarg^{jj'}(\y)
 = 0
\end{multline}
where the $v_i^{jj'}$ are second-order cross-derivatives. 
Collect all these equations in vector form by defining $\a_i(\y)$ as a vector collecting all entries $v_i^j(\y)\, v_i^{j'}(\y), j=1,...,n,j'=1,...,j-1$ (we omit diagonal terms, and by symmetry, take only one half of the indices). Likewise, collect all the entries $v_i^{jj'}(\y), j=1,...,n,j'=1,...,j-1$ in the vector $\b(\y)$, and all the entries $\qmarg^{jj'}(\y), j=1,...,n,j'=1,...,j-1$ in the vector $\cb(\y)$. We can thus write the $n(n-1)/2$ equations above as a single system of equations
\begin{equation} \label{lineqgen}
\sum_i \a_i(\y)  q_i^{11}(v_i(\y),\uu)
+  \b_i(\y)  q_i^{1}(v_i(\y),\uu)
 = \cb(\y)
\end{equation}
Now, collect the $\a$ and $\b$ into a matrix $\M$:
\begin{equation} \label{mdef}
\M(\y)=\begin{pmatrix} \a_1(\y),...,\a_n(\y),\b_1(\y),...,\b_n(\y) \end{pmatrix}
\end{equation}
 Equation (\ref{lineqgen}) takes the form of the following linear system
\begin{equation} \label{matrixconstrgen}
\M(\y) \w(\y,\uu)= \cb(\y)
\end{equation}
where $\w$ is defined in the Assumption of Variability, Eq.~(\ref{wdef}). This must hold for all $\y$ and $\uu$. Note that the size of $\M$ is $n(n-1)/2\times \ncond$.

Now, fix $\y$. Consider the $\ncond+1$ points $\uu_j$ given for that $\y$ by the Assumption of Variability.
Collect the equations (\ref{matrixconstrgen}) above for the $\ncond$ points starting from index $1$:
\begin{equation} \label{mat1gen}
\M(\y)
\begin{pmatrix}
\w(\y,\uu_1),...,\w(\y,\uu_{\ncond})
\end{pmatrix}
 =\begin{pmatrix}
\cb(\y),\ldots,\cb(\y) \end{pmatrix}
\end{equation}
and collect likewise the equation for index $0$ repeated $2n$ times:
\begin{equation} \label{mat2gen}
\M(\y)
\begin{pmatrix}
\w(\y,\uu_0),...,\w(\y,\uu_{0})
\end{pmatrix}
=\begin{pmatrix}\cb(\y),\ldots,\cb(\y) \end{pmatrix}
\end{equation}
Now, subtract (\ref{mat2gen}) from (\ref{mat1gen}) to obtain
\begin{multline} \label{mat3gen}
\M(\y) 
\begin{pmatrix}
\w(\y,\uu_1)-\w(\y,\uu_0),...,\\ \w(\y,\uu_{\ncond}) -\w(\y,\uu_0)
\end{pmatrix}
 =\mathbf{0}
\end{multline}
The matrix consisting of the $\w$ here has, by the Assumption of Variability, linearly independent columns. It is square, of size $\ncond \times \ncond$, so it is invertible. This implies $\M(\y)$ is zero, and thus by definition in (\ref{mdef}), the $\a_i(\y)$ and $\b_i(\y)$ are all zero. 

In particular, $\a_i(\y)$ being zero implies no row of the Jacobian of $\vb$ can have more than one non-zero entry. This holds for any $\y$. 
By continuity of the Jacobian and its invertibility, the non-zero entries in
the Jacobian must be in the same places for all $\y$: If they switched
places, there would have to be a point where the Jacobian is singular,
which would contradict the assumption of invertibility of $\h$.

This means that each $v_i$ is a function of only one $y_i$. The invertibility of $\vv$ also implies that each of these scalar functions is invertible. 
Thus, we have proven the convergence of our method, as well as provided a new identifiability result for nonlinear ICA. 

\section{Proof of Theorem~\ref{Thnew}} \label{proofnew}

For notational simplicity, consider just the case $n=2,k=3$; the results are clearly simple to generalize to any dimensions. Furthermore, we set $Q_i\equiv 1$; again, the proof easily generalizes. 
The assumption of conditional exponentiality means
\begin{align}
q_1(s_1,\uu)=\suffstat_{11}(s_1) \lambda_{11}(\uu) + \suffstat_{12}(s_1) \lambda_{12}(\uu)\nonumber \\+ \suffstat_{13}(s_1) \lambda_{13}(\uu) -\log Z_1(\uu) \\
q_2(s_2,\uu)=\suffstat_{21}(s_2) \lambda_{21}(\uu) + \suffstat_{22}(s_2) \lambda_{22}(\uu) \nonumber\\+ \suffstat_{23}(s_2) \lambda_{23}(\uu) -\log Z_2(\uu)
\end{align}
and by definition of $\w$ in (\ref{wdef}), we get
\begin{multline}
\w(\s,\uu)= \\
\begin{pmatrix}
\suffstat_{11}'(s_1)\lambda_{11}(\uu) + \suffstat_{12}'(s_1)\lambda_{12}(\uu) + \suffstat_{13}'(s_1) \lambda_{13}(\uu) \\
\suffstat_{21}'(s_2)\lambda_{21}(\uu) + \suffstat_{22}'(s_2)\lambda_{22}(\uu) + \suffstat_{23}'(s_2) \lambda_{23}(\uu) \\
\suffstat_{11}''(s_1)\lambda_{11}(\uu) + \suffstat_{12}''(s_1)\lambda_{12}(\uu) + \suffstat_{13}''(s_1) \lambda_{13}(\uu) \\
\suffstat_{21}''(s_2)\lambda_{21}(\uu) + \suffstat_{22}''(s_2)\lambda_{22}(\uu) + \suffstat_{23}''(s_2) \lambda_{23}(\uu) \\
\end{pmatrix}
\end{multline}
Now we fix $\s$ like in the Assumption of Variability, and drop it from the equation. The $\w(\s,\uu)$ above can be written as
\begin{multline}
\begin{pmatrix}
\suffstat_{11}'\\
0\\
\suffstat_{11}''\\
0
\end{pmatrix}
\lambda_{11}(\uu)
+
\begin{pmatrix}
\suffstat_{12}'\\
0\\
\suffstat_{12}''\\
0\\
\end{pmatrix}
\lambda_{12}(\uu)
+
\begin{pmatrix}
\suffstat_{13}'\\
0\\
\suffstat_{13}''\\
0\\
\end{pmatrix}
\lambda_{13}(\uu)
\\+
\begin{pmatrix}
0\\
\suffstat_{21}'\\
0\\
\suffstat_{21}''
\end{pmatrix}
\lambda_{21}(\uu)
+
\begin{pmatrix}
0\\
\suffstat_{22}'\\
0\\
\suffstat_{22}''
\end{pmatrix}
\lambda_{22}(\uu)
+
\begin{pmatrix}
0\\
\suffstat_{23}'\\
0\\
\suffstat_{23}''
\end{pmatrix}
\lambda_{23}(\uu)
\end{multline}
So, we see that $\w(\s,\u)$ for fixed $\s$ is basically given by a linear combination of $nk$ fixed ``basis'' vectors, with the $\lambda$'s giving their coefficients.

If $k=1$, it is impossible to obtain the $\ncond$ linearly independent vectors since there are only $n$ basis vectors. On the other hand, if $k>1$, the vectors $k$ vectors for each $i$ span a 2D subspace by assumption. For different $i$, they are clearly independent since the non-zero entries are in different places. Thus, the $nk$ basis vectors span a $2n$-dimensional subspace, which means we will almost surely obtain $2n$ linearly independent vectors $\w(\s,\uu_i),i=1,\ldots,2n$ by this construction for $\lambda_{ij}$ independently and randomly chosen from a set of non-zero measure (this is a sufficient but by no means a necessary condition).
 Subtraction of $\w(\s,\uu_0)$ does not reduce the independence almost surely, since it is simply redefining the origin, and does not change the linear independence.


\section{Proof of Theorem~\ref{Th2}} \label{proof2}

Denote  
by $\bar{q}_i(s_i)$ the marginal log-density of $s_i$.
As in the proof of Theorem~\ref{Th1}, assuming infinite data, well-known theory says that the regression function will converge to
\begin{multline} 
\sum_{i=1}^n \psi_i(h_i(\x),\uu) =
\log p(\s,\uu) + \log|\J\g(\x) |
- \log p(\s)\\ -\log p(\uu) - \log|\J\g(\x) |\\ 
= \sum_i \log Q_i(s_i) +  [\sum_j \suffstat_{ij}(s_i) \lambda_{ij}(\uu)]  - \log Z_i(\uu) - \qmarr(\s) 
\end{multline}
provided that such a distribution can be approximated by the regression function. Here, we define $\qmarr(\s)= \log p_\s(\s)$. In fact, the approximation is clearly possible since the difference of the log-pdf's is linear in the same sense as the regression function.
In other words, a solution is possible as
\begin{multline} \label{equality}
\sum_{ij} \tilde{h}_{ij}(\x)^T v_{ij}(\uu) + a(\x) + b(\uu) =
\sum_{ij} \suffstat_{ij}(s_i) \lambda_{ij}(\uu) \\+
 \sum_i \log Q_i(s_i)   - \qmarr(\s)
 - \log Z_i(\uu)
\end{multline}
with 
\begin{align}
\tilde{h}_{ij}(\x)&=\suffstat_{ij}(\x)\\
v_{ij}(\uu)&= \lambda_{ij}(\u)\\
a(\x)&= \sum_i \log Q_i(s_i)- \qmarr(\s)\\
b(\u)&= \sum_i - \log Z_i(\uu) 
\end{align}
 Thus, we can have the special form for the regression function in (\ref{rmonoexp}). Next, we have to prove that this is the only solution up to the indeterminacies given in the Theorem.

Collect these equations for all the $\uu_k$ given by Assumption 3 in the Theorem. Denote by $\L$ a matrix of the $\lambda_{ij}(\uu_k)$, with the product of $i,j$ giving row index and $k$ column index. Denote a vector of all the sufficient statistics of all the independent components as  $\mathbf{\suffstat}(\x)=(\suffstat_{11}(s_1),...,\suffstat_{nk}(s_n))^T$.
 Collect all the $\vv(\uu_k)^T$ into a matrix $\V$ with again $k$ as the column index. Collect the terms $ \sum_i \log Z_i(\uu_k) + b(\uu_k)  $ for all the different $k$ into a vector $\z$.

Expressing (\ref{equality}) for all the time points in matrix form, we have
\begin{equation} \label{matrixeq}
\V^T \tilde{\h}(\x) = \L^T \mathbf{\suffstat}(\s) - \z + \mathbf{1} [\sum_i  \log Q_i (s_i) - \qmarr(\s) - a(\x)]  
\end{equation} 
where $\mathbf{1}$ is a $T\times 1$ vector of ones.
Now, on both sides of the equation, subtract the first row from each of the other rows. We get
\begin{equation}
\bar{\V}^T \tilde{\h}(\x) = \bar{\L}^T \mathbf{\suffstat}(\s) - \bar{\z} 
\end{equation} 
where the matrices with bars are such differences of the rows of $\V^T$ and $\L^T$, and likewise for $\z$. We see that the last term in (\ref{matrixeq}) disappears.

Now, the matrix $\bar{\L}$ is indeed the same as in Assumption~3 of the Theorem, which says that the modulations of the distributions of the $s_i$ are independent in the sense that $\bar{\L}$ is invertible. Then, we can multiply both sides by the inverse of $\bar{\L}$ and get
\begin{equation}
\A \tilde{\h}(\x) =\mathbf{\suffstat}(\s)  - \tilde{\z}
\end{equation} 
with an unknown matrix $\A=\bar{\L}^{-1}\bar{\W}$, and a constant vector $\tilde{\z}=\bar{\L}^{-1}\bar{\z}$.

Thus, just like in TCL, we see that the hidden units give the sufficient statistics $\mathbf{\suffstat}(\s)$, up to a linear transformation $\A$, and the Theorem is proven.

\section{Alternative formulation of the Assumption of Variablity} \label{ThApp.sec}
To further strengthen our theory, we provide an alternative formulation of the Assumption of Variability. We define the following alternative:
\begin{enumerate}
\item[] \textbf{[Alternative Assumption of Variability]} Assume $\uu$ is continuous-valued, and that there exist $\ncond$ values for $\uu$, denoted by $\uu_j,j=1...\ncond$ such that the $\ncond$ vectors in $\Rb^{2n}$ given by
\begin{equation}
(\walt(\y,\uu_1),\walt(\y,\uu_2),...,\walt(\y,\uu_{\ncond}))
\end{equation}
with
\begin{multline}
\walt(\s,\uu)=(\frac{\partial^2 q_1(s_1,\uu)}{ \partial s_1 \partial u_j},\ldots,\frac{\partial^2 q_n(s_n,\uu)}{ \partial s_n  \partial u_j},\\  \frac{\partial^3 q_1(s_1,\uu)}{ \partial s_1^2 \partial u_j},\ldots,\frac{\partial^3 q_n(s_n,\uu)}{ \partial s_n^2 \partial u_j})
\end{multline}
are linearly independent, for some choice of the auxiliary variable index $j$.  \label{assvaralt}
\end{enumerate}
Theorem~\ref{Th1} holds with with this alternative assumption as well. In the proof of the Theorem, take derivatives of both sides of (\ref{mat1gen}) with respect to the $u_j$ in the Theorem. Then, the right-hand-side vanishes, and we have an equation similar to (\ref{mat1gen}) but with $\walt$. All the logic after (\ref{mat3gen}) applies to that equation.

\section{Using a function of $\x$ as auxiliary variable} \label{ufx.sec}

We provide an informal proof without full generality to show why defining $\u$ as a direct deterministic function of $\x$ is likely to violate the assumption of conditional independent. Consider 
   a simple linear mixing $x_1=s_1+s_2$ (with something similar for
   $x_2$), and define tentatively $u=x_1$. Conditioning $s_1$ on
   $u$ will now create the dependence $s_1=x_1-s_2=u-s_2$ which violates
   conditional independence. (This example would be more realistic
   with additive noise $u=x_1+n$ to avoid degenerate pdf's, but the same logic
   applies anyway.) In fact, if we could make the model identifiable
   by such $\u$ defined as a function of $\x$, we would have violated the
   basic unidentifiability theory by Darmois. Thus, conditional
   independence implies that $\u$ must bring new information in
   addition to $\x$, and this information must be, in some very loose
   intuitive sense, "sufficiently independent" of the information in $\x$.

\section{Additional discussion to Section~\ref{pclcomp.sec}} \label{pclcompappendix.sec}

In \citep{Hyva17AISTATS}, the model was proven to be identifiable under two assumptions: First, the joint log-pdf of two consecutive time points is not ``factorizable'' in the conditionally exponential form of order one,
A variant of such dependency was called ``quasi-Gaussianity'' in \citep{Hyva17AISTATS}. However, here we use a different terminology to highlight the connection to the exponential family important in our theory as well as TCL. There is also a slight difference between the two definitions, since in \citep{Hyva17AISTATS}, it was only necessary to exclude the case where the two functions in the factorization are equal, i.e.\ $\suffstat_1=\lambda_1$ in the current notation.
The second assumption was that there is a rather strong kind of temporal dependency between the time points, which was called uniform dependency. Here, we need no such latter condition, essentially because here we constrain $\h$ to be invertible, which was not done in \citep{Hyva17AISTATS}, but seems to have a somewhat similar effect.

\section{Additional discussion to Section~\ref{combine.sec}} \label{combineappendix.sec}

One might ask whether it would better to randomize $t$ and $\x(t-1)$ separately, by using two independent random indices $t^*$ and $\x(t^{**}-1)$. The choice between these two should be made based on how to modulate the conditional distribution $p(s_i|t,\x(t-1))$ as strongly as possible. In practice, we would intuitively assume it is usually best to use a single time index as above, because then the dependency in $t^*$ and $\x(t^*-1)$ will make the modulation stronger. 
Moreover, the Theorems above would not apply directly to a case where we have two different random indices, although the results might be easy to reformulate for such a case as well.

\section{Acknowledgments}

A.H.\ was supported by CIFAR and the Gatsby Charitable Foundation.
H.S.\ was supported by JSPS \text{KAKENHI} 18K18107.
R.E.T.\ thanks EPSRC grant EP/M026957/1.

\end{document}